\documentclass[a4paper]{article}

\pdfoutput=1
\usepackage{hyperref}
\hypersetup{
 pdfinfo={
   Title={Generating Material Maps to Map Informal Settlements},
   Author={Patrick Helber, Bradley Gram-Hansen, Indhu Varatharajan, Faiza Azam, Alejandro Coca-Castro, Veronika Kopackova, Piotr Bilinski}
 }
}

\PassOptionsToPackage{numbers, compress}{natbib}
\usepackage[final]{nips_2018}

\usepackage[utf8]{inputenc} 
\usepackage[T1]{fontenc}    
\usepackage{hyperref}       
\usepackage{url}            
\usepackage{booktabs}       
\usepackage{amsfonts}       
\usepackage{nicefrac}       
\usepackage{microtype}      
\usepackage{graphicx}  
\usepackage[final]{pdfpages}
\usepackage{subfig}
\usepackage{wrapfig}
\usepackage[polish,english]{babel}
\title{Generating Material Maps to Map \\ Informal Settlements}

\author{Patrick Helber\thanks{authors contributed equally} \\
     DFKI and TU Kaiserlautern \\
     \texttt{\href{mailto:patrick.helber@dfki.de}{\small{patrick.helber@dfki.de}}} \\
    \And 
   Bradley Gram-Hansen{$^*$}\\
     University of Oxford \\
     \texttt{\href{mailto:bradley@robots.ox.ac.uk}{\small{bradley@robots.ox.ac.uk}}}\\
    \And
    Indhu Varatharajan \\
     DLR\\
     \texttt{\href{mailto:indhu.varatharajan@dlr.de}{\small{indhu.varatharajan@dlr.de}}}\\
    \And
    Faiza Azam \\
     Independent researcher \\
     \texttt{\href{mailto:fazam@hotmail.de}{\small{fazam@hotmail.de}}}\\
    \And 
    Alejandro Coca-Castro \\
     King's College London \\
     \texttt{\href{mailto:alejandro.coca\textunderscore castro@kcl.ac.uk}{\small{alejandro.coca\_castro@kcl.ac.uk}}}\\
         \And 
    Veronika Kopackova \\
     Czech Geological Survey \\
     \texttt{\href{mailto:veronika.kopackova@seznam.cz}{\small{veronika.kopackova@seznam.cz}}} \\
         \And 
    Piotr Biliński \\
     University of Oxford and University of Warsaw \\ 
     \texttt{\href{mailto:piotrb@robots.ox.ac.uk}{\small{piotrb@robots.ox.ac.uk}}}\\
     }
     
\begin{document}
\maketitle

\begin{abstract}


Detecting and mapping informal settlements encompasses several of the United Nations sustainable development goals. This is because informal settlements are home to the most socially and economically vulnerable people on the planet. Thus, understanding where these settlements are is of paramount importance to both government and non-government organizations (NGOs), such as the United Nations Children’s Fund (UNICEF), who can use this information to deliver effective social and economic aid. We propose a method that detects and maps the locations of informal settlements using only freely available, Sentinel-2 low-resolution satellite spectral data and socio-economic data. This is in contrast to previous studies that only use costly very-high resolution (VHR) satellite and aerial imagery. We show how we can detect informal settlements by combining both domain knowledge and machine learning techniques,  to build a classifier that looks for known roofing materials used in informal settlements. Please find additional material at \mbox{\url{https://frontierdevelopmentlab.github.io/informal-settlements/}}.
\end{abstract}

\section{Informal Settlements}

The United Nations~(UN) and the Organisation for Economic Co-operation and Development~(OECD) state that informal settlements are defined as follows~\cite{2008oecd,united2012state}: 
\begin{quote}
\begin{enumerate}
\item Inhabitants have no security of tenure vis-\`a-vis the land or
dwellings they inhabit, with modalities ranging from squatting to informal rental housing. 
\item The neighborhoods usually lack, or are cut off from, basic services and city infrastructure.
\item The housing may not comply with current planning and building regulations, and is often situated in geographically and environmentally hazardous areas.
\end{enumerate}
\end{quote}

Typically, those that live in informal settlements are the most vulnerable in society, subject to harsh social and economic constraints~\cite{WEKESA2011238}. Although informal settlements are well studied in the humanities and remote sensing communities \cite{united2012state,huchzermeyer2006informal,hofmann2008detecting}, in machine learning little has been done to map informal settlements using high-resolution satellite imagery~\cite{mahabir2018critical,mboga2017detection,varshney2015targeting} and as far as we know nothing has been done using low-resolution imagery.   Being able to map and locate these settlements would give organizations such as UNICEF and other UN organizations the ability to provide effective social and economic aid~\cite{pais2002poverty}.

\paragraph{Data Sources}

\begin{wrapfigure}[16]{r}{0.35\textwidth}
\vspace*{0.4cm}
  \includegraphics[width=0.49\linewidth]{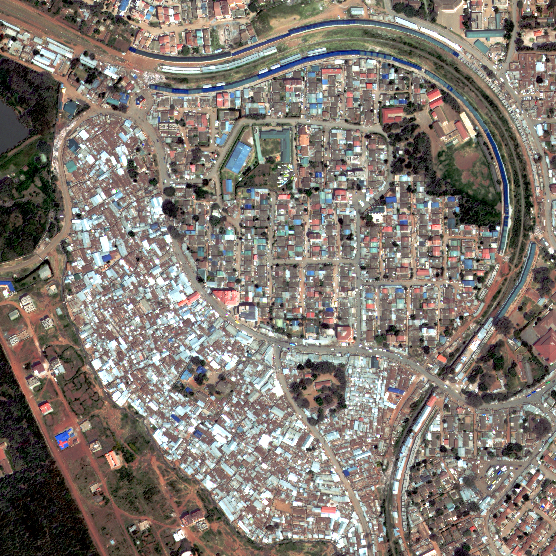}\hfill
  \includegraphics[width=0.49\linewidth]{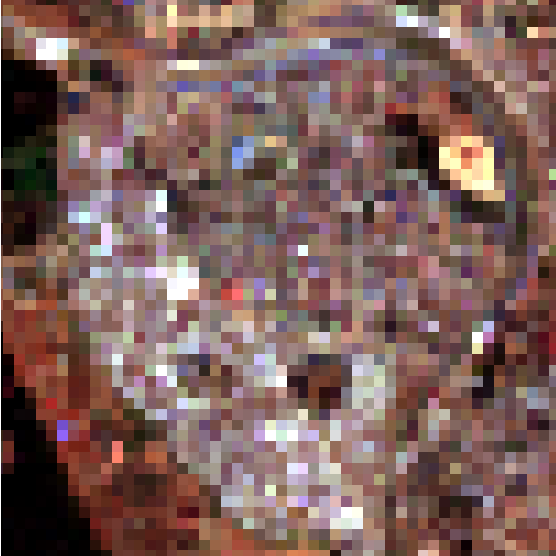}

\caption{\footnotesize Two images of the same informal settlement in Kibera, representing the difference between high and low resolution imagery. \textit{Left}: A DigitalGlobe 30$cm$ resolution VHR image, meaning each pixel represents a 30$cm^2$ area. \textit{Right}: The Sentinel-2 10$m$ resolution image, that is, each pixel represents a  10$m^2$ area.}
\label{fig:compresimg}
\end{wrapfigure}

\paragraph{Low-resolution Sentinel-2 Satellite Data} Each pixel in a Sentinel-2 image contains a thirteen dimensional feature vector.  This feature vector includes, besides the usual RGB bands, additional ten bands that are acquired at different wavelengths throughout the visible (VIS) and near-infrared (NIR) spectral range. These pixels are of a 10$m$ resolution, which means that each pixel represents a $10m\times10m$ surface. Thus, there is a lot of contextual information contained within one pixel. By observing the spectral signal, which provides us with the chemical composition of a pixel, we can extract this contextual information. See Figure~\ref{fig:compresimg}. 

\begin{wrapfigure}[18]{r}{0.5\textwidth}
\vspace*{-0.4cm}
\centering
    \includegraphics[width=0.245\linewidth]{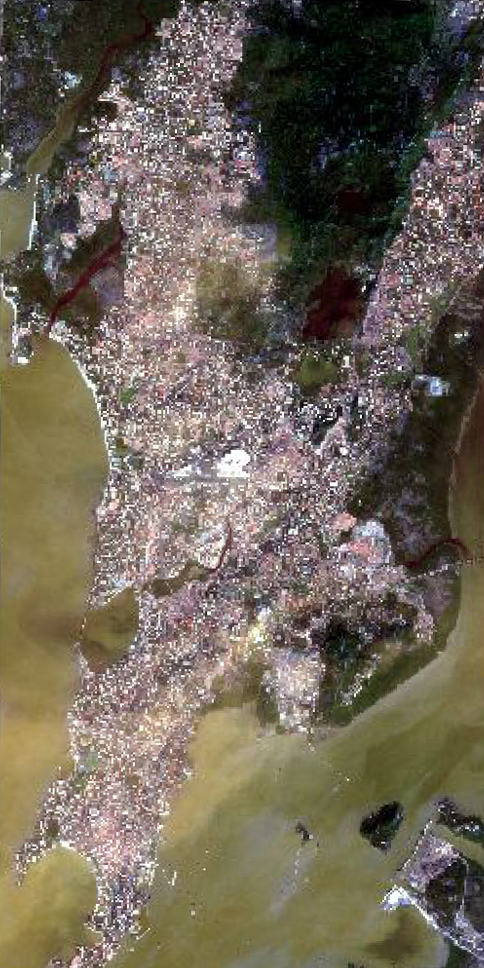}
    \includegraphics[width=0.245\linewidth]{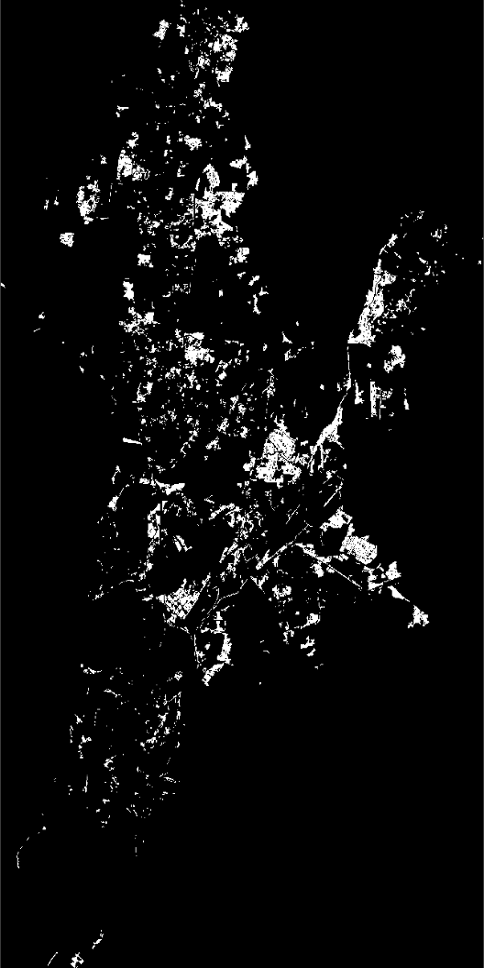}
    \includegraphics[width=0.245\linewidth]{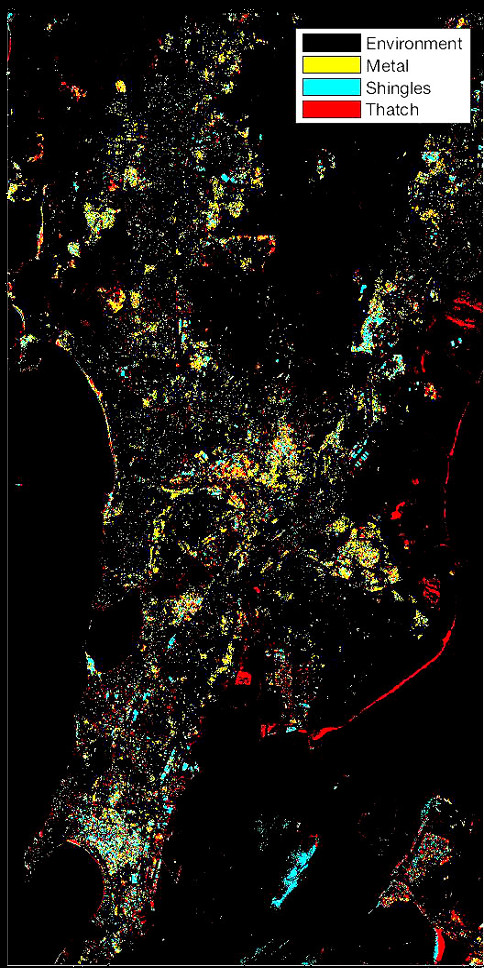}
   \caption{\footnotesize{Material prediction of Mumbai with \mbox{$n_{trees}{=}10$}. \textit{Left:} Sentinel-2 image. The area is densely populated and contains several informal settlements. \textit{Middle:} Partially complete ground truth for the Informal settlements, white represents informal settlements and black represents environment. The Southern right hand corner is not included in this mask. \textit{Right:}  A CCF prediction on the Sentinel-2 L1C spectral data for determining what materials are present. Black is environment, yellow is metal, blue is shingles and red is thatch.}}
   \label{fig:mumbaimat}
\end{wrapfigure}

\paragraph{Afrobarometer Data}
We take advantage of rounds 1-6 of the open source database Afrobarometer~\cite{Afro}, which is a pan-African non-partisan research network that conducts socio-economic public surveys serving policy making.  In partnership with US Global Development Lab, AIDdata, the Afrobarometer survey are mapped to specific villages and towns in 36 African nations providing hyperlocal time-series information. In particular, we have used Afrobarometer data as ground truth, as the dataset contains geo-located survey data, which asked \emph{What was the roof of the respondent’s home or shelter made of?} Respondents could choose from the following options: 1) Metal, tin or zinc, 2) Tiles, 3) Shingles, 4) Thatch or grass, 5) Plastic sheets, 6) Asbestos,
7) Multiple materials, 8) Some other material, 9) Could not tell/could not see. 
It is important to note, that whilst this data should be seen as ground truth, it itself is noisy. This is due to two reasons. 1) Even though the data is geo-located the location provided did not, on multiple occasions, align with any type of roof, or material other than vegetation. 2) Distortion in the geo-located coordinates is added to protect the privacy of the respondents. Because of this we use several pre-processing steps to remove any noisy data points, which reduced the number of classes to four.

\section{Method}
\label{sec:method}

Our approach uses domain knowledge about the types of materials used to build informal settlements. The freely available Afrobarometer~\cite{Afro} data and low-resolution Sentinel-2~\cite{Copernicussent2} satellite imagery is used to train a classifier to detect those type of materials.

\paragraph{Canonical Correlation Forests}~(CCFs)~\cite{rainforth2015canonical} are a decision tree ensemble method for classification and regression.
CCFs are the state-of-the-art random forest technique, which have shown to achieve remarkable results for numerous regression and classification tasks~\cite{rainforth2015canonical}.
Individual canonical correlation trees are binary decision trees with hyperplane splits based on local canonical correlation coefficients calculated during training. Like most random forest based approaches, CCFs have very few hyper-parameters to tune and typically provide very good performance out of the box. The only parameter that has to be set is the number of trees, $n_{trees}$. For CCFs, setting $n_{trees} = 15$ provides a performance that is empirically equivalent to a random forest that has $n_{trees} = 500$~\cite{rainforth2015canonical}, meaning CCFs have lower computational costs, whilst providing better classification.
CCFs work by using canonical correlation analysis~(CCA) and projection bootstrapping during the training of each tree, which projects the data into a space that maximally correlates the inputs with the outputs. This is particularly useful when we have small datasets, like in our case, as it reduces the amount of artificial randomness required to be added during the tree training procedure and improves the ensemble predictive performance~\cite{rainforth2015canonical}.\\

 The computational efficiency aspects of CCFs and their suitability to both small and large datasets, makes them ideal for detecting informal settlements for three reasons. First, many of the organisations that we aim to help will not have access to a large amount of compute resources, therefore computational efficiency is important. Second, to run the CCFs for both training and prediction, all that has to be called is one function. This ensures that the end user does not need to be an expert in ensemble methods and makes the method akin to plug and play. Finally, our ground truth datasets are relatively small, for example, in the Afrobarometer data we have 11 data points per class for training. This means that we have to use the data as efficiently as possible, which CCFs allow us to do.

\section{Results}

\begin{wrapfigure}[32]{r}{0.45\textwidth}
\vspace*{-0.4cm}
\centering
    \includegraphics[width=.42\textwidth]{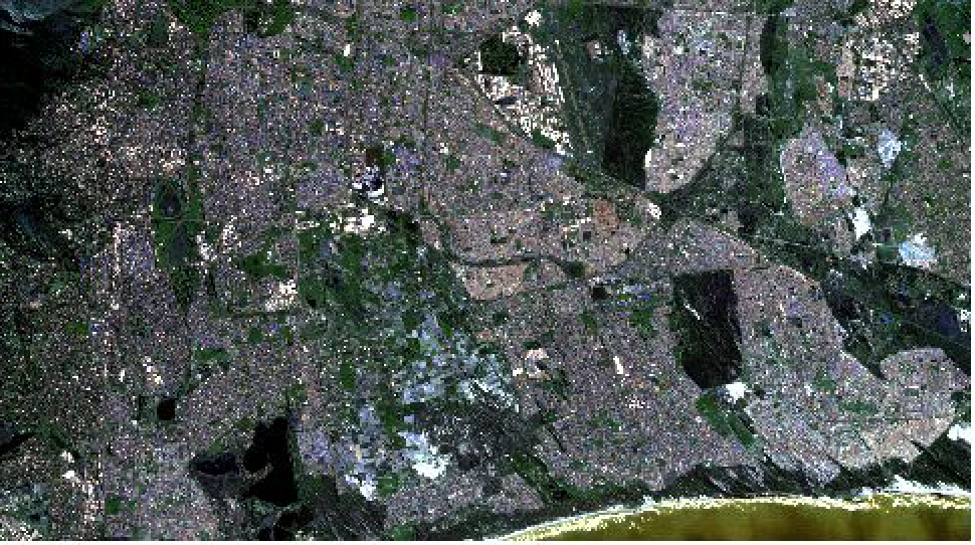} \hfill
    \includegraphics[width=.42\textwidth]{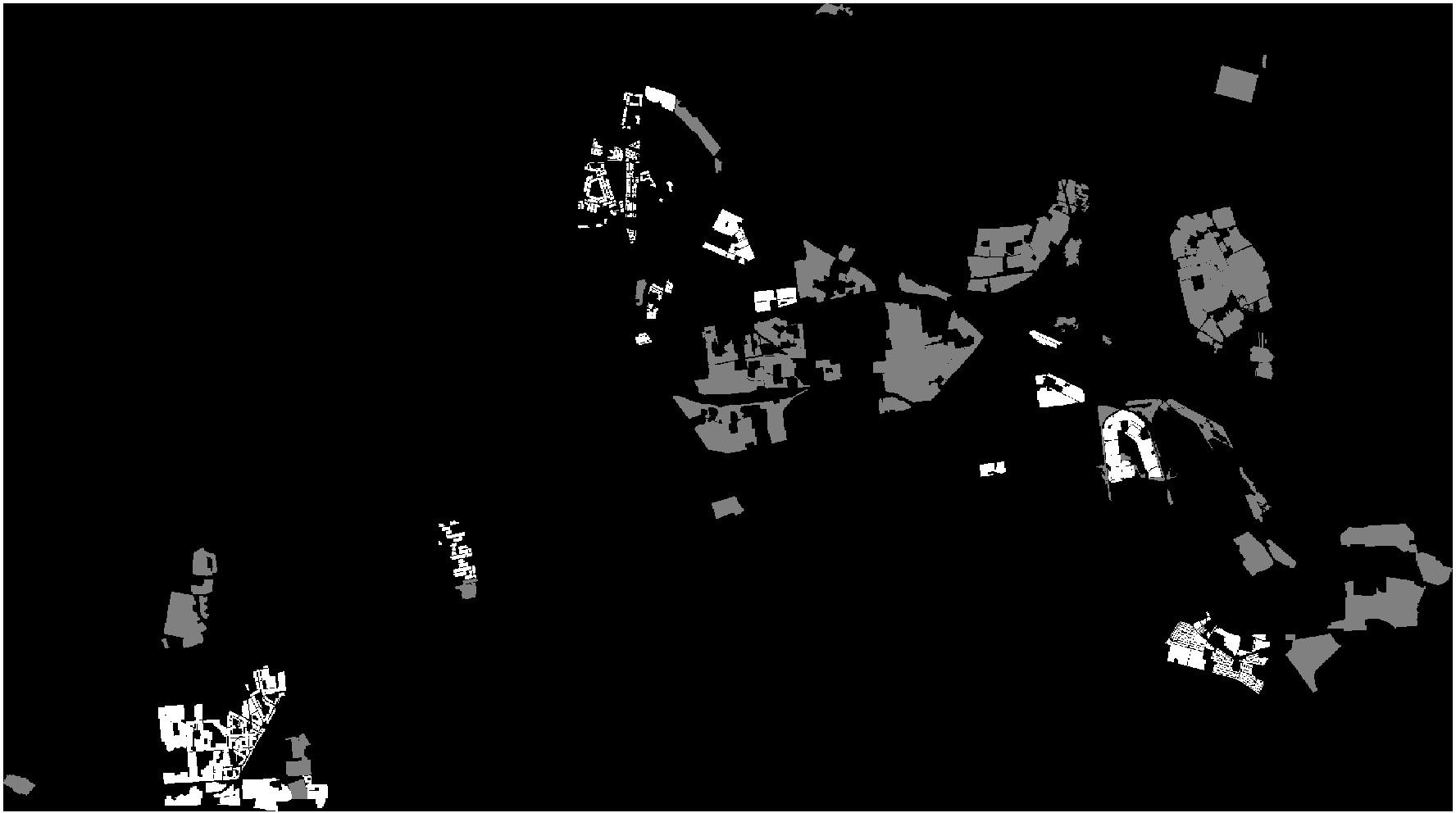}\hfill
    \includegraphics[width=.42\textwidth]{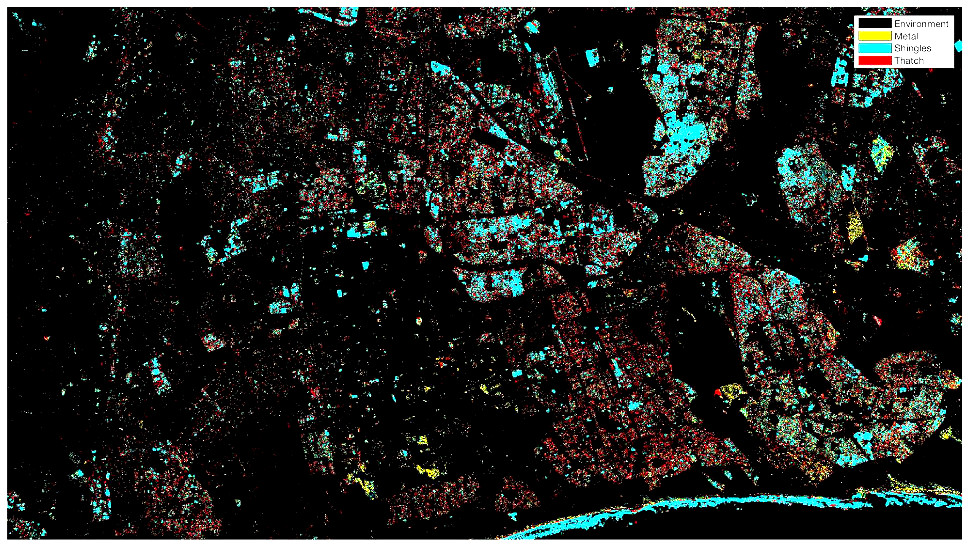}
  \caption{\footnotesize Material prediction of Captetown with $n_{tees}{=}10$. \textit{Top:} Sentinel-2 image of Capetown. \textit{Middle:} Ground truth is 35\% complete. White represents an informal settlement and black represents environment. \textit{Bottom:} A CCF material prediction on the Sentinel-2 L1C spectral data.}
\label{fig:capetwn}
\end{wrapfigure}
\paragraph{Experimental Setup}
The classes that we choose to predict are the metal, shingles, thatch and environment class. Metal contains aluminum, zinc and tin signals. The shingles class contains asbestos, some types of metal and wood shingles. The thatch class contains a spectrum that is similar to that of dry vegetation and the environment class contains everything else.

We train only on the Sentinel-2 spectral signal extracted from the geo-located Afrobarometer data points regarding respondents roof type. Due to the inconsistency of the data, we have 11 Level-1C product spectral signals per class for training to ensure that the training data is balanced, as the largest class contains 373 annotations, whereas the lowest contained just 16 annotations. In order to validate our models we have to rely on the help of domain experts to verify our predictions, as we have no ground truth annotations regarding the spectral material of each pixel in our test images.

As can be seen in Figures \ref{fig:mumbaimat} and \ref{fig:capetwn}, the material prediction heuristically provides a useful model for detecting informal settlements within Mumbai and Cape Town by looking for known materials used in the construction of informal settlements. There are some clear mis-classifications, however, we hope that if we were to have larger amounts of training data, future models could be made more robust. It should also be noted that the ground truth maps are not entirely complete, especially for Cape Town. In order to develop a more formal way to compare results we are working with multiple partners to develop a larger dataset that consists of Sentinel-2 Level-1C and Level-2A spectral signals, with the corresponding material annotations. This would enable us to evaluate our predictions much more robustly without the need of an external expert.

\paragraph*{Future Work}
Currently much of the work performed in this area requires multiple partnerships over varying disciplines and institutions, which makes it difficult to conduct research effectively. We are currently working on ways to make it easier for the machine learning community to participate within this area and related areas, by constructing datasets and highlighting socio-economic problems that need to be solved. 

\paragraph*{Acknowledgments.}
This project was executed during the Frontier Development Lab~(FDL), Europe program, a partnership between the Phi-Lab at ESA, the Satellite Applications~(SA) Catapult, Nvidia Corporation, Oxford University and Kellogg College. We gratefully acknowledge the support of Adrien Muller and Tom Jones of SA Catapult for their useful comments, providing VHR imagery and ground truth annotations for Nairobi. We thank UNICEF, in particular Do-Hyung Kim and Clara Palau Montava, for valuable discussions and AIData for access to geo-located Afrobarometer data. We thank Nvidia for computation resources. We thank Yarin Gal for his helpful comments. Patrick Helber was supported by the NVIDIA AI Lab program and the BMBF project DeFuseNN (Grant 01IW17002). Bradley Gram-Hansen was also supported by the UK EPSRC CDT in Autonomous Intelligent Machines and Systems.

\bibliographystyle{plain}
\bibliography{refs}
\end{document}